\documentclass[runningheads]{llncs}
\usepackage{hyperref}
\usepackage[T1]{fontenc}
\usepackage{amssymb}
\usepackage{amsmath}
\usepackage{cleveref}
\usepackage{xcolor}
\usepackage{graphicx}

\begin{document}

\title{Terrain-Aware Thompson Sampling for Multi-Drone Search and Rescue}

\author{Riana Gagnon Souleiman\inst{1} \and Giovanni Beltrame\inst{2}}

\authorrunning{Gagnon \& Beltrame}

\institute{Polytechnique Montreal, Montreal QC H3T0A3, Canada}

\maketitle              
\begin{abstract}
In this work, we propose a decentralized terrain-aware multi-drone Thompson Sampling algorithm tailored for search and rescue (SAR) applications. A probability distribution of locating a lost person is informed by terrain data, using established SAR models (displacement ring, elevation, watershed, dispersion angle, and track offset) derived from historical SAR case data of lost person behavior. Exploration is conducted by a team of drones using Thompson Sampling, where candidate actions are selected by drawing samples from this probability distribution (a per-cell Gaussian posterior), whose mean and variance are updated through Bayesian inference. This approach enables a balance between exploration of uncertain regions and exploitation of high-probability areas with the goal of locating a lost person. Observation uncertainty is incorporated through fractional forest coverage, measuring the percent of an observation occluded by tree coverage and accounting for reduced detection performance in densely vegetated areas. Simulations are performed using real SAR case data (YOSAR) to evaluate the effectiveness of the proposed method compared to a coverage planner and probabilistic active search method in a 1.5 $km^2$ and 3.8 $km^2$ search region with 5,10,15, and 20 agents flying at 50 km/hr. We demonstrate that our method is able to successfully locate a lost person at a success rate of ${\sim}87\%$ within $60$--$247$~s for an area of 1.5~$km^2$, and at a success rate of ${\sim}75-91\%$ within $161$--$466$~s for an area of 3.8~$km^2$, across configurations.

\end{abstract}
\section{Introduction}

Wilderness search and rescue (WiSAR) operations are inherently arduous, time-consuming, and resource-intensive, requiring careful planning and efficient execution to maximize the probability of survival of a lost person. In land-based SAR missions, survivability decreases rapidly over time, making timely and informed decision-making critical. To support this process, a variety of models have been developed \cite{Koester2008,Doke2012AnalysisOS,mapscore} to generate probability maps that estimate the likely locations of a missing individual. These approaches are grounded in mathematical search theory \cite{Koopman1980}, which addresses the fundamental challenge of locating an unknown target within a large search space. While such probabilistic approaches significantly improve search efficiency, SAR operations still demand extensive human effort, coordination, and time to achieve successful outcomes.

Multi-robot systems, in particular teams of unmanned aerial vehicles (UAVs), offer a promising solution by enabling rapid, parallel exploration of large areas \cite{Baxter2007,Drew2021SARReview,SARSurvey}. However, the effectiveness of such systems depends critically on their ability to operate autonomously and in a distributed manner. Centralized coordination approaches reduce robustness in SAR scenarios due to limited communication infrastructure, bandwidth constraints, and the risk of single points of failure. To address these challenges, decentralized task allocation methods have been proposed, enabling each agent to independently and cooperatively make decisions while contributing to a shared objective \cite{trajenpy_book,TANG2018132,disttask}.

Thompson Sampling (TS), a Bayesian decision-making framework increasingly used in search and rescue \cite{thompsonsearch,activethompson,distthompson}, provides a natural basis for decentralized decision-making under uncertainty. TS maintains a probabilistic belief over unknown quantities via a prior probability distribution that is updated as observations are collected (posterior). Agents sample from this posterior to select actions with the highest sampled reward, naturally balancing exploration and exploitation. When each agent maintains its own posterior and shares observations with agents within communication range, this enables a scalable, parallel decision-making process without a central coordinator.

We propose a decentralized, terrain-aware Thompson sampling framework for multi-drone SAR applications that incorporates key terrain characteristics such as linear features, elevation, and watershed structures to inform prior probability distributions based on established models of lost person behavior. Each UAV maintains a local belief over the search space and selects actions by sampling from this distribution over candidate actions. The proposed framework is evaluated in simulation using real-world SAR case data (YOSAR\footnote{https://hub.arcgis.com/maps/620a96d2429249a187bdc4a7cc058cf5/about}), with the objective of locating a missing individual in large, remote environments. Overall, this work aims to improve the scalability, robustness, and efficiency of multi-drone SAR operations.

\textbf{In summary, our major contributions are:}
\begin{enumerate}
    \item A decentralized terrain-aware Thompson sampling algorithm that leverages existing search and rescue models to assign informed priors for search of lost persons in large, remote environments.
    \item Object detection uncertainty based on fractional forest coverage.
    \item Experimental simulations analyzing the performance of the proposed method using real search and rescue cases.
\end{enumerate}

\section{Related Works}
Decentralized methods for multi-agent search and rescue (SAR) offer several desirable properties, including robustness, scalability, flexibility, and reduced communication requirements~\cite{MRTA}. By eliminating reliance on a central coordinator, these approaches can effectively operate over large areas with large teams while avoiding single points of failure. A significant portion of work in multi-drone SAR has focused on coverage-based planning, which aims to ensure complete and exhaustive exploration of the search area~\cite{spantree,rlcover,coopplan,linuav}, typically generating trajectories that evenly distribute agents across the environment. Grøntved et al.~\cite{grontved2022icar} proposed a decentralized task allocation framework based on the consensus-based bundle algorithm (CBBA), enabling drones to coordinate trajectory assignments for coverage path planning. While such approaches achieve performance comparable to centralized methods, they primarily emphasize area coverage and do not explicitly optimize for rapid target discovery. In contrast, our approach prioritizes high-probability regions through terrain-informed priors, enabling faster target localization rather than uniform exploration.

To improve search efficiency for search and rescue applications, probabilistic methods have been explored. Ewers et al. introduced predictive probability mapping techniques that model the likely behavior of missing persons using real-world SAR data ~\cite{Ewers_2025,Ewers_2023}. These methods generate probability distributions over the search space to guide human responders; however, they do not incorporate active search by autonomous agents. Our method builds on these probabilistic priors but integrates them directly into multi-robot decision-making framework, enabling a team of autonomous agents to actively explore large remote areas efficiently.

Active search strategies based on Thompson sampling have demonstrated strong performance in balancing exploration and exploitation when locating objects of interest under uncertainty~\cite{ghods2021,nats}. Bakshi et al. proposed the GUTS framework, which leverages uncertainty-aware Thompson sampling for multi-agent coordination ~\cite{guts}. Their method allocates tasks to heterogeneous agents by sampling from a sparse posterior distribution, enforced through a zero-mean prior. While effective in promoting exploration, this approach does not incorporate domain-specific priors, such as typical missing person behaviors, which can be critical in SAR applications. In contrast, our framework incorporates terrain-informed priors to prioritize high-probability regions in the search area while incorporating inter-agent repulsion to reduce redundant exploration for efficient and coordinated search in realistic SAR environments.

\section{Methodology}

\subsection{Problem Formulation}
Given a discretized search space $\mathbf{M} = \{c_i\}_{i=1}^{|\mathbf{M}|}$, where each $c_i$ denotes a grid cell, the objective is to locate a missing person as quickly as possible. The unknown target location is represented by a sparse binary vector $\boldsymbol{\beta} \in \{0,1\}^{|\mathbf{M}|}$, where $\beta_i = 1$ indicates the presence of the target in cell $c_i$. Each cell $c_i$ is associated with an initial prior belief modeled as a Gaussian distribution, $p_0(\beta_i) = \mathcal{N}(\mu_{c_i}, 1)$ where $\mu_{c_i}$ is derived from terrain-informed SAR priors (see \cref{sec:priors}). The observation uncertainty is defined as $\sigma^2_{c_i} \propto \, o_{c_i} $ where $o_{c_i}$ denotes the fractional forest coverage.

Each agent $j$ operates over a local planning region $\mathbf{M}^j_L \subseteq \mathbf{M}$, defined as a fixed-size bounded neighborhood consisting of a $10 \times 10$ grid window centered at the agent’s current position in the global search space. A trajectory for agent $j$ is defined as a sequence of visited cells, $\tau^j = (c^j_i)_{i=1}^{S}, \quad c^j_i \in \mathbf{M_L^j}$ where a target cell $c^j_S$ is selected using Thompson sampling by drawing samples from the current belief over the local region and selecting the cell with the highest sampled value. As the agent executes its trajectory, it collects observations, incorporated through Bayesian updates to refine the posterior belief over $\boldsymbol{\beta}$. After executing a trajectory, agents share their observations with the team, and the planning process is repeated under the updated belief until the target is located or mission constraints are met. We assume continuous inter-agent communication (i.e., global connectivity) and impose a battery constraint of 1800 seconds. All agents are initialized at a common starting location corresponding to cell $c_0$.

\subsection{Map Priors}\label{sec:priors}
 The map $\mathbf{M}$ is discretized into $30\,\text{m} \times 30\,\text{m}$ grid cells corresponding to the sensing area of a drone flying at an altitude of 80 meters with a vertical and horizontal camera field of view of $21^\circ$.
 Each cell is assigned a probability value computed as the average weighted sum of probability density functions from multiple search and rescue models. Track-offset, watershed, and elevation models assign probabilities based on environmental features, while displacement-ring and dispersion angle depend on the last known location (LKL) and intended destination. The displacement-ring model decreases probability with distance from the LKL, while the dispersion angle model increases likelihood along feasible directions between the LKL and destination.

\begin{figure}[htpb]
\includegraphics[width=\textwidth]{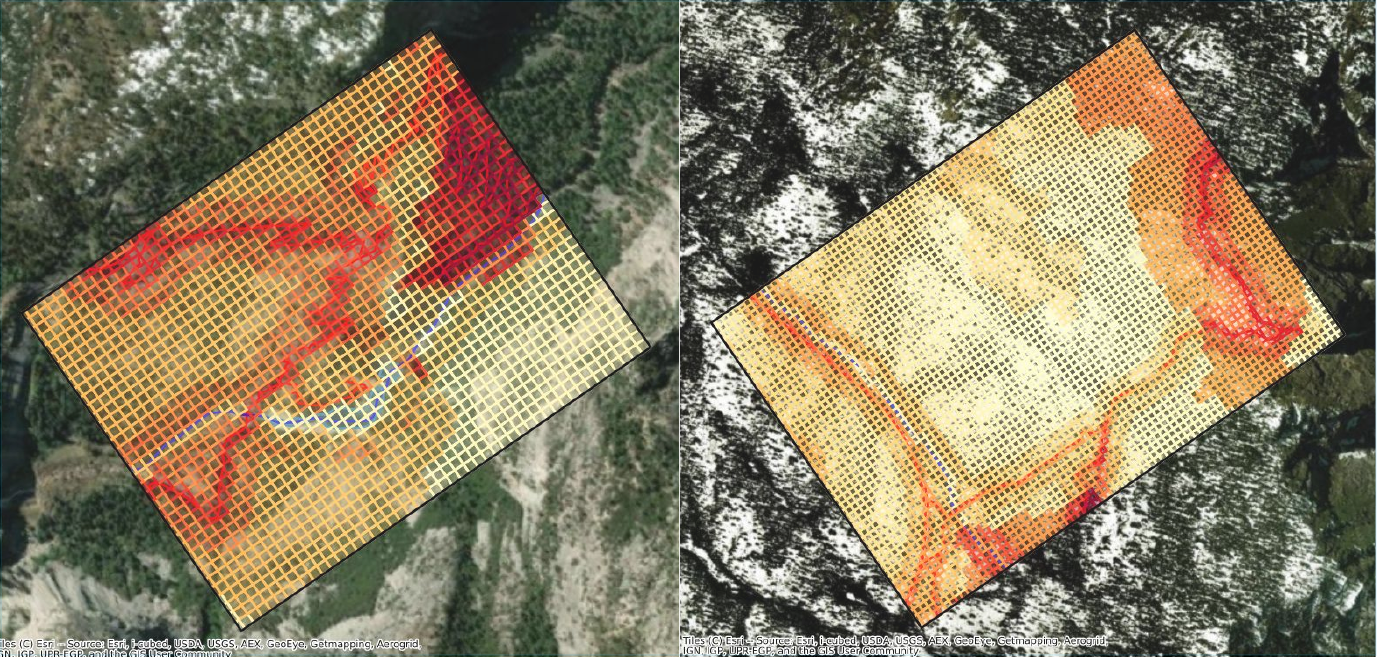}
\caption{Prior mean values over two search areas, where each grid cell represents a 30~m $\times$ 30~m physical area. Cell color encodes the prior probability of the missing person's location, scaled from yellow (low probability) to red (high probability). Left: 1.5~km$^2$ and Right: 3.8~km$^2$ search areas, with the number of grid cells scaling with area.} \label{fig:POA}
\end{figure}

Linear features such as roads and waterways are extracted using \cite{grontved2022icar}, informing the track-offset model by assigning higher likelihood near these features and decaying with distance. Elevation is derived from NASA SRTM data, allowing computation of relative elevation with respect to the LKL and identification of uphill and downhill regions. Watershed delineation is performed using \cite{bartos_2020}, reducing likelihood across watershed boundaries relative to the LKL. The resulting combined probability distribution over all cells is visualized in \cref{fig:POA} for two representative search areas.

\begin{figure}[b]
\includegraphics[width=\textwidth]{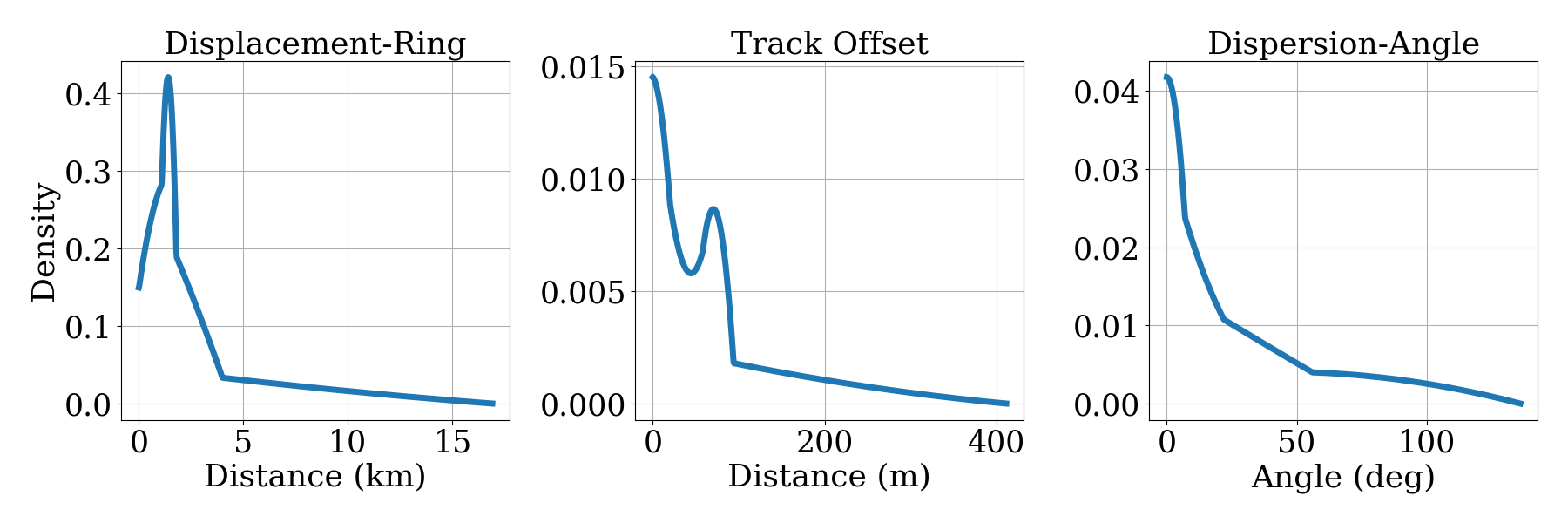}
\caption{Resulting Probability Density Functions (PDFs) for displacement-ring (left), track offset (middle) and dispersion-angle (right) used in calculation of map priors.} \label{fig:PDFs}
\end{figure}

The mean probability value $\mu$ for each grid cell $c_i \in \mathbf{M}$ is defined as a normalized weighted mixture of the $K$ probability density functions $f_k(\cdot)$ derived from the heterogeneous search features introduced above (terrain, distance, heading angle). Each $f_k(c_i) \geq 0$ is the unnormalized contribution of the $k$-th model evaluated at cell $c_i$. The normalized mixture is
\begin{equation}\label{eq:pdfs}
\mu_{c_i} =
\frac{\sum_{k=1}^{K} \alpha_k f_k(c_i)}
{\sum_{c_i' \in \mathbf{M}} \sum_{k=1}^{K} \alpha_k f_k(c_i')}\,,
\end{equation}
where $\alpha_k \geq 0$ are non-negative weighting coefficients set to $\alpha_k = 0.2$ for all $K = 5$ models to weight each SAR submodel equally, which we found to be a reasonable default in the absence of a learned combination. The denominator normalizes the mixture so that $\sum_{c_i \in \mathbf{M}} \mu_{c_i} = 1$, yielding a valid discrete probability distribution over the search space.

Each $f_k$ is obtained from historical SAR data. For the dispersion-angle model, we interpolate an empirical cumulative distribution $F_k(\theta) = \mathrm{Interp}(\theta; \{\theta_i, F_i\})$ over observed heading angles and differentiate to recover a continuous density $f_k(\theta) = \tfrac{d}{d\theta}F_k(\theta)$ (\cref{fig:PDFs} right). The remaining models are constructed analogously over their respective features. Non-negativity and normalization of $\mu_{c_i}$ then follow immediately from $f_k(c_i) \geq 0$, $\alpha_k \geq 0$, and the denominator above.

\subsection{Multi Agent Search}
\begin{figure}[b]\label{fig:flowchart}
\includegraphics[width=\textwidth]{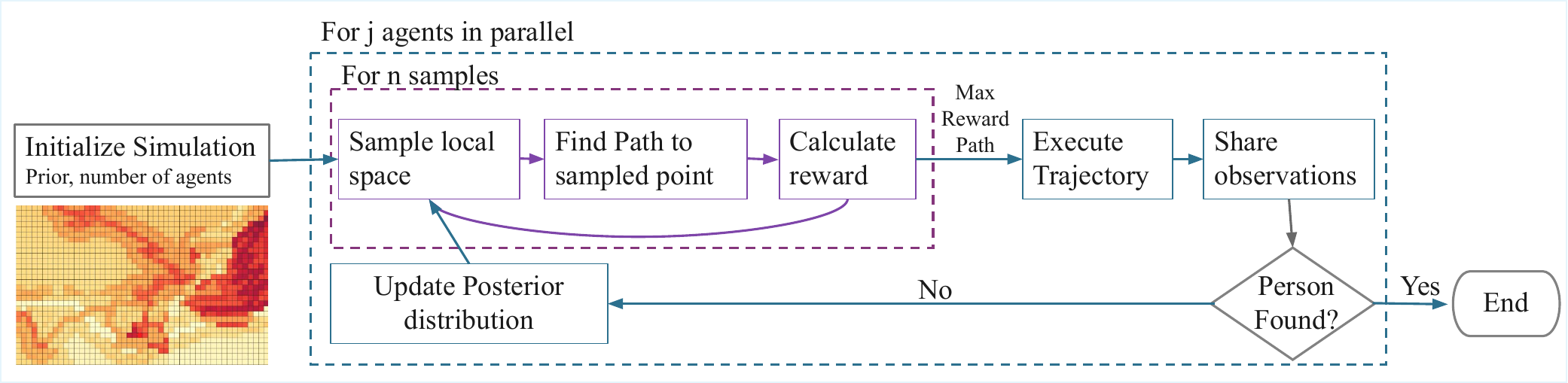}
\caption{State machine of the proposed multi-agent active search algorithm. All agents execute the algorithm in parallel. Agents sample candidate locations, plans paths to each, and evaluate a reward function over n iterations. The best trajectory is executed and observations are shared to other agents. If the missing person is located, the search terminates; otherwise, the posterior belief map is updated and the process repeats.}
\end{figure}

Each agent, $j$, selects an action corresponding to a trajectory $\tau^j \in \mathcal{T}$ from a set of feasible trajectories evaluated under the agent’s current belief. All agents begin at a common initial location $c_0$ with a shared prior belief over the search space where $p_0(\beta_{c_i}) = \mathcal{N}(\mu_{c_i},\gamma_{c_i})$, where $\mu_{c_i}$ is derived in \cref{sec:priors} and $\gamma_{c_i}$ is the variance initialized as 1. This work follows the method outlined in \cite{guts,nats,ghods} for updating the posterior, which computationally optimizes parallelized uncertainty-aware Thompson Sampling and utilizes expectation maximization to estimate a posterior distribution $p(\boldsymbol{\beta}|D_t^j,\Gamma)$ where $D^j_t$ is the set of observations at timestep $t$ of robot $j$ and $\Gamma = \mathrm{diag}([\gamma_1\dots\gamma_{|\mathbf{M}|}])$. We introduce a diagonal observation-weighting matrix $W$, with $W_{tt} = 1 - \lambda\sigma_{c(t)}^2$, where $c(t)$ is the cell sensed at time $t$ and $\lambda \in [0,1]$ is the scaling factor used in \cref{prop:occlusion} to control how strongly occlusion suppresses an observation. The posterior, $p(\boldsymbol{\beta}|D^j_t,\Gamma)$, can then be estimated as $\mathcal{N}(\boldsymbol{\mu},\mathbf{V})$ where:
\begin{equation}\label{eq:std}
    \mathbf{V} = (\Gamma^{-1} + \mathbf{X}^T W \mathbf{X})^{-1}
\end{equation}
\begin{equation}\label{eq:mean}
\boldsymbol{\mu} = \mathbf{V}\left(\Gamma^{-1}\boldsymbol{\mu}_0 + \mathbf{X}^T W \mathbf{y}\right)
\end{equation}
\begin{equation}\label{gamma}
    \gamma_{c_i} = ([V]_{c_ic_i}+[\mu]^2_{c_i}+2)/(1+2a)
\end{equation}

Here $\mathbf{X} \in \{0,1\}^{T \times |\mathbf{M}|}$ is the design matrix whose $t$-th row $\mathbf{X}_t$ is the one-hot indicator of the cell sensed at time $t$, $\mathbf{y} \in \{0,1\}^{T}$ stacks the corresponding binary observations (1 indicating the lost person), and $W \in \mathbb{R}^{T \times T}$ is the diagonal observation-weight matrix defined above; with these conventions $\mathbf{X}^{T} W \mathbf{X}$ is an $|\mathbf{M}| \times |\mathbf{M}|$ diagonal matrix and $\mathbf{X}^{T} W \mathbf{y}$ is a vector of length $|\mathbf{M}|$. The pair $(\mathbf{X}, \mathbf{y})$ collects the measurements in $D^j_t$ shared among all agents, $\boldsymbol{\mu}_0$ is the mean at the previous planning step, and $a$ is a constant set to 0.1. An overview of the algorithm workflow is provided in \cref{fig:flowchart}.

\subsubsection{Uncertainty}
Spatial perception uncertainty is directly related to the degree of occlusion, where higher occlusion corresponds to increased uncertainty \cite{perceptionuncert}. For each cell, the uncertainty $\sigma_{c_i}^2 \in [0,1]$ encodes the level of ground occlusion (\%), obtained from fractional forest coverage data from Google Earth Engine. To account for occlusion, we weight updates such that high-uncertainty measurements contribute less to belief updates. However, since $\sigma_{c_i}^2 \in [0,1]$ can excessively suppress updates in highly occluded regions, we introduce a scaling factor $\lambda$ to regulate the effective update strength and ensure stable convergence of the inference process. We set $\lambda = 0.665$, which we found to provide a stable update rate.

\begin{proposition}[Occlusion-weighted convergence rate]\label{prop:occlusion}
Let $w_{c_i}^{(t)} = 1 - \lambda\sigma_{c_i}^2$ with $\lambda \in [0,1]$, and let $W_{c_i}^{(T)} = \sum_{t=1}^{T} w_{c_i}^{(t)}$ denote the cumulative observation weight accrued at cell $c_i$ after $T$ updates. Under the one-hot design $\mathbf{X}$ used in \cref{eq:std,eq:mean}, the posterior variance and the posterior mean at cell $c_i$ satisfy
\begin{equation}\label{eq:conv_rates}
V_{c_i}^{(T)} = \left(\gamma_{c_i}^{-1} + W_{c_i}^{(T)}\right)^{-1} = O\!\left(\big(W_{c_i}^{(T)}\big)^{-1}\right)\textbf{,} \ \mu_{c_i}^{(T)} - \mu_{c_i}^{(\infty)} = O\!\left(\big(W_{c_i}^{(T)}\big)^{-1/2}\right)
\end{equation}
so that heavily occluded cells (where $\sigma_{c_i}^2 \to 1$ and $w_{c_i} \to 0$) see slower belief convergence, whereas lightly occluded cells converge more quickly. The scaling factor $\lambda$ bounds the update rate and prevents excessively slow convergence that would cause agents to cluster in a limited region.
\end{proposition}

\paragraph{Derivation.}
Because $\mathbf{X}$ is one-hot, $\mathbf{X}^{T} W \mathbf{X}$ is diagonal and its $c_i$-th diagonal entry equals $\sum_{t=1}^{T} w_{c_i}^{(t)} = W_{c_i}^{(T)}$. Substituting into \cref{eq:std} yields the scalar form $V_{c_i}^{(T)} = (\gamma_{c_i}^{-1} + W_{c_i}^{(T)})^{-1}$, from which the $O((W_{c_i}^{(T)})^{-1})$ variance rate follows directly. The posterior mean in \cref{eq:mean} is a precision-weighted average of the prior mean and the observed values with combined precision $\gamma_{c_i}^{-1} + W_{c_i}^{(T)}$; its residual with respect to the large-sample limit is therefore bounded by $\sqrt{V_{c_i}^{(T)}}$, yielding the $O((W_{c_i}^{(T)})^{-1/2})$ rate for the mean. Finally, substituting $w_{c_i}^{(t)} = 1 - \lambda \sigma_{c_i}^2$ ties each increment of $W_{c_i}^{(T)}$ directly to the per-cell occlusion, so higher occlusion reduces increments and slows convergence at both stated rates.

\subsubsection{Action Selection}
At each decision step, samples are drawn from the posterior distribution, and the action with the highest reward $\mathcal{R}$ is selected. By sampling from a normal distribution, $\mathcal{N}(\mu,V)$, the method naturally balances exploration and exploitation. While all agents share observations over a common global map, decision-making is performed independently on each agent without a central coordinator: agent $j$ maintains its own copy of the posterior belief and samples action candidates over $\mathbf{M}_L^j$. As a result, agents may assign similar high rewards to the same regions, leading to redundant exploration behavior. To mitigate this effect, a repulsion term is introduced into the reward function to discourage agents from selecting the same targets, thereby promoting spatial diversity and improving overall search efficiency. The repulsion factor is defined as:

\begin{equation}
U_{\text{repulse}}(c_i) =
\left\{
    \begin{array}{lr}
        \frac{\eta}{2}\left(\frac{1}{d_{c,c^j}}-\frac{1}{d_0}\right)^2, & \text{if } d_{c,c^j}\leq d_0\\
        0, & \text{if } d_{c,c^j} > d_0
    \end{array}
\right.
\end{equation}

where $\eta$ is a scaling factor set to $\eta = 1000$, ensuring that the resulting values are of a similar magnitude to the sampled posterior value, $d_{c,c^j}$ is the distance between the position of a cell along the trajectory and the position of a neighboring agent, and $d_0$ is a 300 m radius around the agent. The reward function is then defined as the sum of the sampled observations $\tilde{z}_{c_i}$, minus the repulsion cost over all cells along the trajectory $\tau^j$:
\begin{equation}
    \mathcal{R}(\tau^j) = \sum_{c_i \in \tau^j} \left( \tilde{z}_{c_i} - U_{\text{repulse}}(c_i) \right)
\end{equation}
such that $\tilde{z}_{c_i} \sim \mathcal{N}(\mu_{c_i}, \gamma_{c_i})$ given by the posterior distribution. Our trajectory $\tau^j$ is generated using Dijkstra's algorithm to minimize the cumulative path cost:%

\begin{equation}\label{eq:dijkstra_traj}
\hat{\tau}_{0:S} = \arg\min_{\tau} \sum_{i=1}^{S} \left( w \, d(c_i, c_{i-1}) - \tilde{z}_{c_i}  \right)
\end{equation}

Here, $d(c_i, c_{i-1})$ denotes the distance between consecutive cells and $w$ is a weighting factor balancing path length and sampled observation. In practice we choose $w$ large enough that the per-edge cost $w\, d(c_i, c_{i-1}) - \tilde{z}_{c_i}$ remains non-negative for every feasible pair of neighbouring cells, which guarantees the correctness of Dijkstra's shortest-path computation under the bounded range of the sampled posterior values $\tilde{z}_{c_i}$ and the non-negativity of $U_{\text{repulse}}$. This method promotes slight detours toward highly rewarding cells along the trajectory.

\section{Experiments}

\begin{figure}[b]
\includegraphics[width=\textwidth]{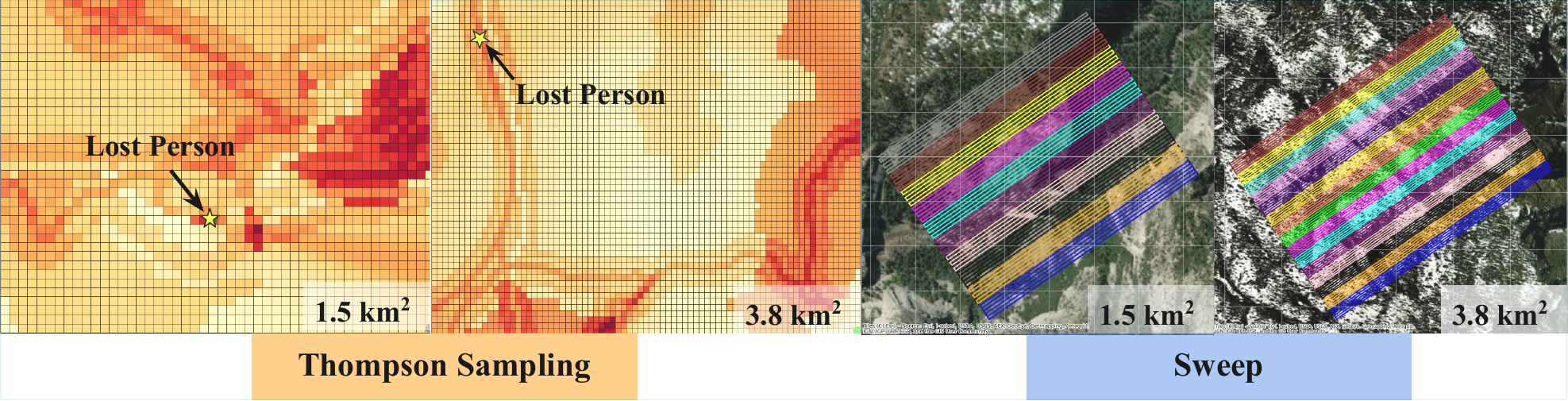}
\caption{Initialization of our proposed Thompson sampling method (left) using terrain-informed priors for different search regions and sweep method (right) with evenly partitioned trajectories among 10 agents in $1.5\,\text{km}^2$ and 15 agents in $3.8\,\text{km}^2$ area with different colors representing individual agent trajectories.} \label{fig:initializesim}
\end{figure}

The proposed method was evaluated through simulation to assess its performance across varying team sizes and search area scales. Each configuration was repeated over $10$ runs per YOSAR case, yielding $30$ trials per configuration across the three cases used:
\begin{enumerate}
    \item \textbf{Number of agents:} [5, 10, 15, 20]
    \item \textbf{Search area:} [1.5, 3.8] $km^2$
\end{enumerate}

Three real-world SAR cases from the Yosemite Search and Rescue (YOSAR) Missing Person Database were used to construct the prior distributions over each search space, incorporating information such as the last known location, intended destination of the missing person and ground truth target locations within the simulated environments. The probability of detection at the target cell was approximately 40\% across all cases, with tree cover data obtained from Google Earth Engine.

The method is compared to two methods: a coverage-based strategy where the search area is evenly partitioned among agents \cite{grontved2022icar}, and GUTS \cite{guts}, an uncertainty-aware Thompson sampling-based active search method that employs a zero-mean prior. Furthermore, an ablation test was performed with a uniform prior to analyze the importance of the terrain-aware prior. It is assumed that agents remain within communication range and fly at a constant speed of 50 $\frac{km}{h}$, and each mission is constrained to a maximum duration of 1800 seconds, corresponding to the typical operational battery life of a drone. A table of all parameter values used in simulation is reported in \cref{tab:params}.
\begin{table}[b]
\centering
\begin{tabular}{@{\hspace{.6cm}}c@{\hspace{.6cm}}|@{\hspace{.6cm}}c@{\hspace{.6cm}}|@{\hspace{.6cm}}c@{\hspace{.6cm}}|@{\hspace{.6cm}}c@{\hspace{.6cm}}|@{\hspace{.6cm}}c@{\hspace{.6cm}}|@{\hspace{.6cm}}c@{\hspace{.6cm}}}
\hline\hline
$\mathbf{\alpha_k}$ & $\mathbf{a}$ & $\mathbf{\eta}$ & $\mathbf{w}$ & $\mathbf{d_0}$ & $\mathbf{\lambda}$ \\
\hline\hline
$0.2$ & $0.1$ & $1000$ & $4.0$ & $300\,\text{m}$ & $0.665$ \\[1.5ex]
\hline\hline
\end{tabular}
\caption{Table of Parameters Used for Simulation}
\label{tab:params}
\end{table}

\subsection{Results}

\begin{figure}[t]
\centering
\includegraphics[width=.9\textwidth]{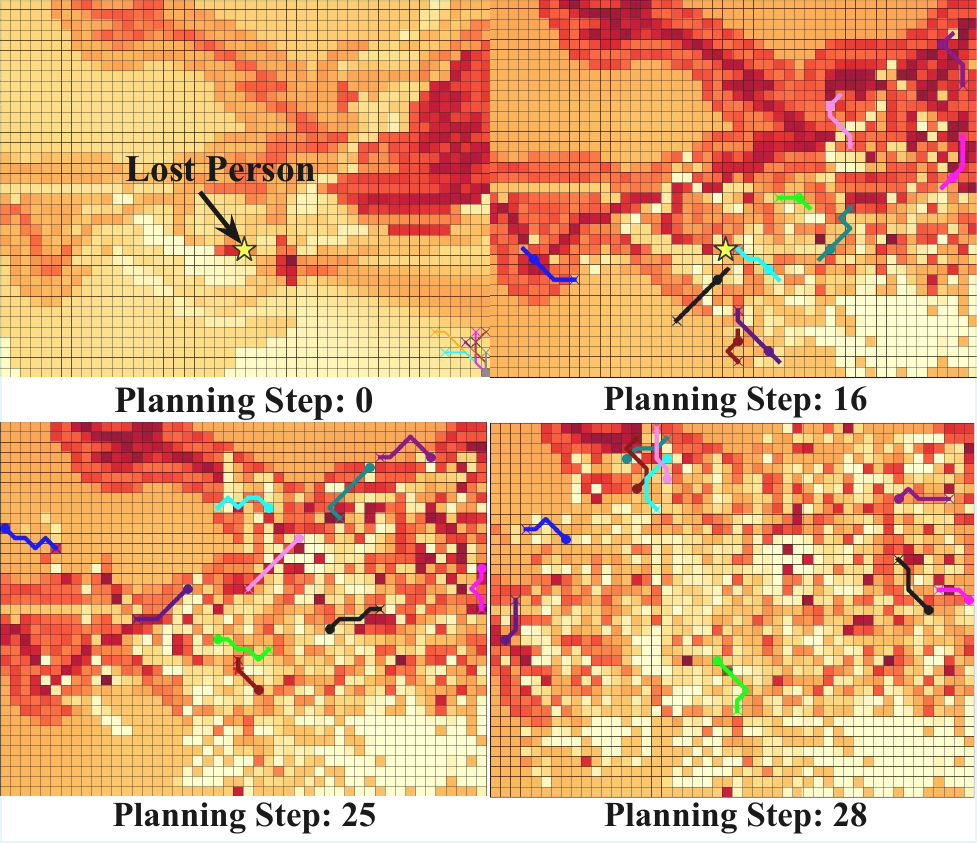}
\caption{Evolution of search for 10 agents in a 1.5 $km^2$ area, plotted over the probability distribution of the missing person's location, scaled from yellow (low probability) to red (high probability) with probability of detecting the target at is 40 \% due to occlusion. Planning step 25 shows an agent visiting the target cell but failing to detect the person due to low sensing probability, while planning step 28 shows successful re-visitation and detection of the target (green star).} \label{fig:evolution}
\end{figure}

\begin{figure}[t]
\includegraphics[width=\textwidth]{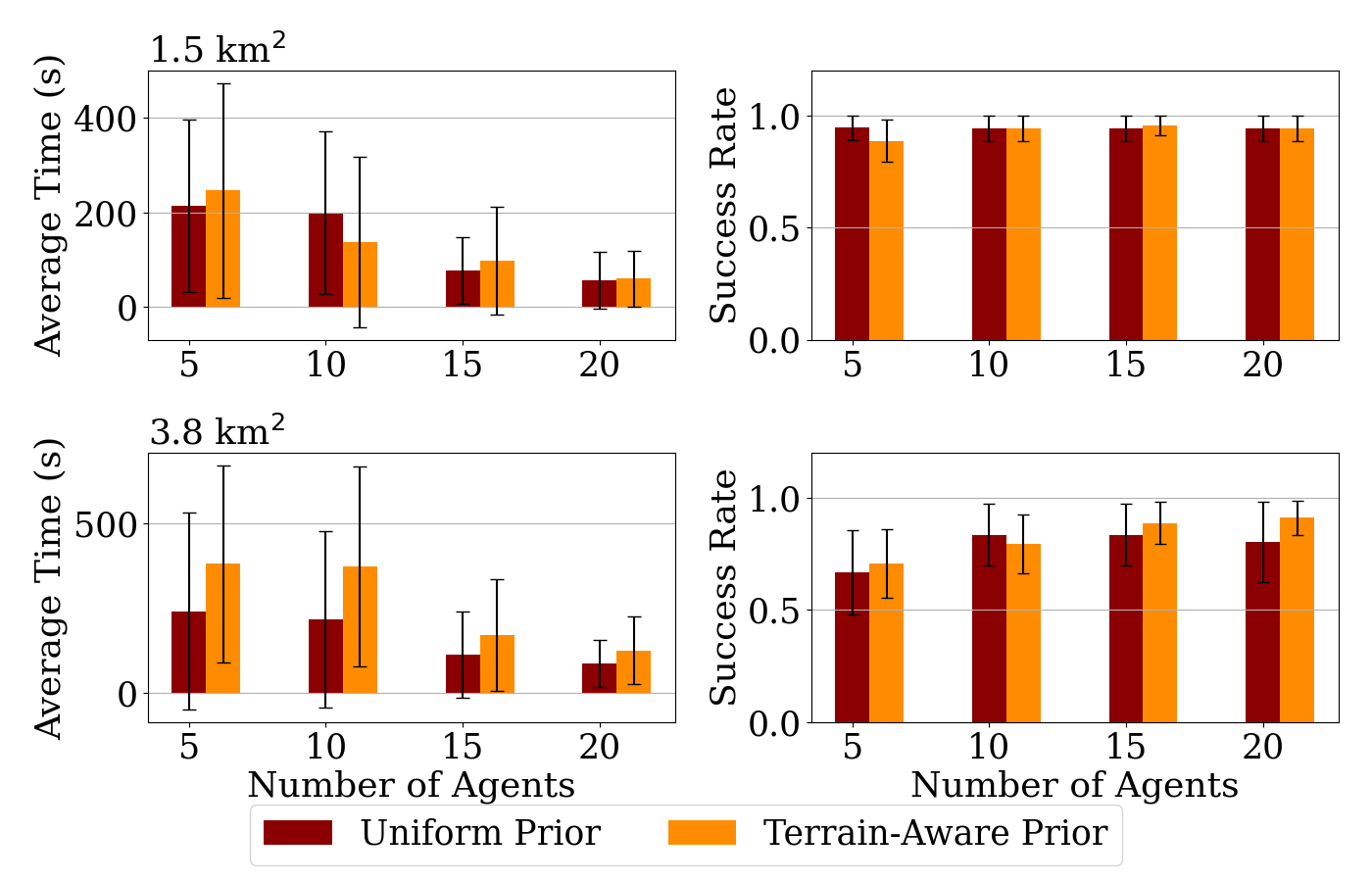}
\caption{Mean Time to Locate (left, mean ± std) and Success Rate (right, 95\% Wilson confidence intervals) over 30 runs of the proposed method with uniform priors (red) and Terrain-Aware priors (orange) in 1.5 $km^2$ (top) and 3.8 $km^2$ (bottom) search regions with 40\% detection probability.} \label{fig:priornoprior}
\end{figure}

\begin{figure}[t]
\includegraphics[width=\textwidth]{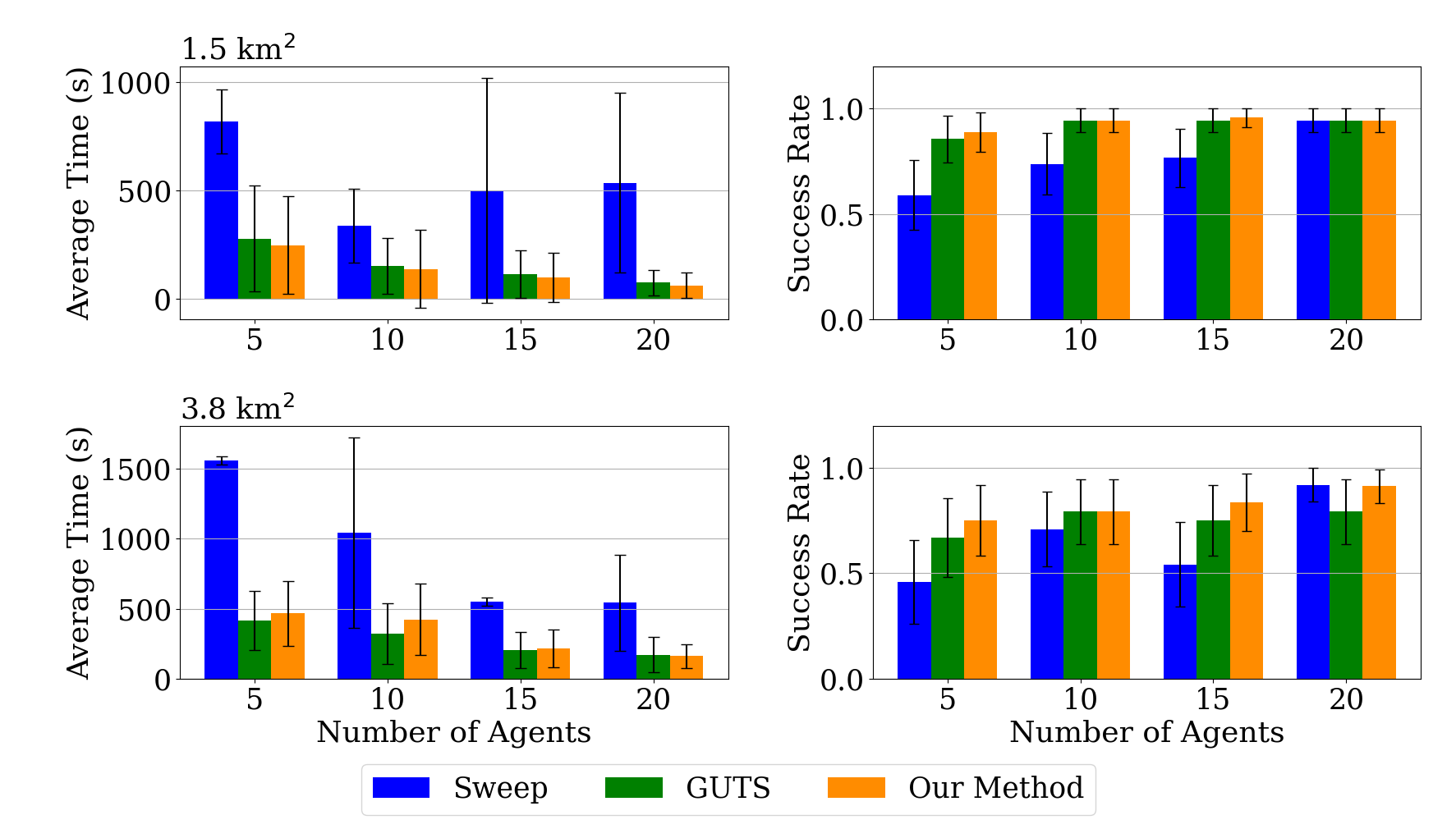}
\caption{Mean Time to Locate (left, mean ± std) and Success Rate (right, 95\% Wilson confidence intervals) over 30 runs for sweep (blue), GUTS (green), and proposed method (orange) in $1.5~km^2$ (top) and $3.8~km^2$ (bottom) search regions with 40\% detection probability.} \label{fig:data}
\end{figure}

The results of this work demonstrate the effectiveness of the proposed method for locating a lost person in large, highly occluded environments using terrain and uncertainty aware Thompson sampling. The occlusion weighting mechanism slows the rate of convergence for cells with high uncertainty, ensuring they remain viable candidates for continued exploration, enabling agents to dynamically return to regions with high variance during search. In the proposed framework, all drones actively explore the environment while maintaining repulsive interactions to reduce overlap and improve coverage efficiency, incorporating observation uncertainty to preserve potential detections while leveraging terrain-based priors to prioritize high-probability regions of the search space (\cref{fig:evolution}). 

The proposed method was analyzed with both terrain-aware and uniform priors ($\mu = 0)$ to asses the importance of informed priors (\cref{fig:priornoprior}). Across experiments, the terrain-aware prior demonstrates improved robustness in more complex environments compared to a uniform prior. In the larger \(3.8~\text{km}^2\) setting, it consistently achieves equal or higher success rates, with the gap becoming more pronounced as the number of agents increases. This suggests that the terrain-aware prior scales more effectively with additional agents, maintaining or improving performance where the uniform prior begins to plateau or degrade. While detection times are often longer or comparable, they remain competitive overall, with the approach providing improved reliability in more challenging conditions, becoming more pronounced as environment complexity increases.

Furthermore the performance of the proposed method was compared to systematic sweep and GUTs as shown in \cref{fig:data}. Thompson sampling consistently outperforms sweep in time-to-detection across all configurations, while also achieving higher success rates within the 1800-second time budget. Coverage-based methods are inherently disadvantaged in uncertain environments, as they uniformly traverse the search space. If the target is not detected, the entire trajectory is repeated, leading to increased detection times in highly occluded regions. In contrast, Thompson sampling methods avoids this failure mode by maintaining an uncertainty-aware belief, enabling reallocation of visited cells. Results show that detection time decreases as the number of agents increases. In the $1.5~\text{km}^2$ area, our method achieves a mean detection time of $246.4$~s with 5 agents, which decreases to $137.1$~s with 10 agents and falls below $100$~s for 15--20 agents, while maintaining a success rate of ${\sim}87\%$--$96\%$ across all configurations and matching or exceeding GUTS in both detection time and success rate at every agent count. In the larger $3.8~\text{km}^2$ area our method remains effective despite the increased search space, achieving detection times of $466.3$~s and $422.2$~s for 5 and 10 agents, respectively, before dropping significantly to $215.8$~s and $161.3$~s for 15 and 20 agents. In the larger setting, our method offers a higher or comparable success rates, increasing from $75\%$ at 5 agents to $91\%$ at 20 agents. 

Terrain-aware priors therefore provide comparable time-to-detection as uniform priors but excel in larger environments with higher success rates of locating the target. Detection times remain substantially lower than those of sweep-based approaches in every configuration, confirming that Thompson sampling with terrain-aware priors is a practical and scalable strategy for search and rescue applications.

\section{Conclusion}

We present a decentralized terrain-aware multi-agent active search method for search and rescue applications using terrain-aware Thompson sampling with uncertainty. Samples on a posterior distribution informed through existing search and rescue models balance exploration and exploitation to promote active search in high probability areas with sensing uncertainty employed through fractional forest coverage of the observed cell. Results demonstrate that the proposed method is able to locate a lost person at a success rate of ${\sim}87\%$ within $60$--$247$~s for an area of 1.5~$km^2$, and at a success rate of ${\sim}75-91\%$ within $161$--$466$~s for an area of 3.8~$km^2$, across 5--20 agents.

In future work, we aim to relax several simplifying assumptions to better reflect real-world SAR deployments. In particular, we will incorporate communication constraints into the framework, as the assumption of constant connectivity is unrealistic in large or obstructed environments. To address this, a virtual stigmergy~\cite{virtualstyg} will be integrated to enable asynchronous information sharing, allowing observations to propagate across agents despite intermittent communication. Furthermore, the scalability is constrained by the matrix operations; as the area and number of agents increase, so does the number of grid cells and observations, leading to significant computational overhead. We aim to optimize subsequent iterations to improve overall scalability. Finally, we plan to validate the proposed method on physical drone platforms, where real-world factors such as sensing noise, localization error, and energy limitations will provide further insight into the practical applicability of the approach.

\bibliographystyle{plain} 
\bibliography{dars}

@inproceedings{guts,
   title={GUTS: Generalized Uncertainty-Aware Thompson Sampling for Multi-Agent Active Search},
   url={http://dx.doi.org/10.1109/ICRA48891.2023.10160597},
   DOI={10.1109/icra48891.2023.10160597},
   booktitle={2023 IEEE International Conference on Robotics and Automation (ICRA)},
   publisher={IEEE},
   author={Bakshi, Nikhil Angad and Gupta, Tejus and Ghods, Ramina and Schneider, Jeff},
   year={2023},
   month=may, pages={7735–7741} }

@misc{bartos_2020,
    title  = {pysheds: simple and fast watershed delineation in python},
    author = {Bartos, Matt},
    url    = {https://github.com/mdbartos/pysheds},
    year   = {2020},
    doi    = {10.5281/zenodo.3822494}
}

@InProceedings{trajenpy_book,
author="Gr{\o}ntved, Kasper A. R.
and Bahodi, Maria-Theresa
and Christensen, Anders Lyhne",
editor="Marreiros, Goreti
and Grande, Laura
and Llerena, Juan Pedro
and Concei{\c{c}}{\~a}o, Lu{\'i}s
and Ko, Hoon
and Plaza, Marta
and Ricca, Michela",
title="Automated Task Generation for Multi-drone Search and Rescue Operations",
booktitle="Distributed Computing and Artificial Intelligence, Special Sessions II, 21st International Conference",
year="2025",
publisher="Springer Nature Switzerland",
address="Cham",
pages="266--271",
isbn="978-3-031-80946-0"
}

@article{Drew2021SARReview,
  title   = {Multi-Agent Systems for Search and Rescue Applications},
  author  = {Drew, Daniel S.},
  journal = {Current Robotics Reports},
  year    = {2021},
  volume  = {2},
  number  = {2},
  pages   = {189--200},
  month   = jun,
  doi     = {10.1007/s43154-021-00048-3},
  url     = {https://doi.org/10.1007/s43154-021-00048-3},
  issn    = {2662-4087}
}

@Inbook{Baxter2007,
author="Baxter, J. L.
and Burke, E. K.
and Garibaldi, J. M.
and Norman, M.",
editor="Mukhopadhyay, Subhas Chandra
and Gupta, Gourab Sen",
title="Multi-Robot Search and Rescue: A Potential Field Based Approach",
bookTitle="Autonomous Robots and Agents",
year="2007",
publisher="Springer Berlin Heidelberg",
address="Berlin, Heidelberg",
pages="9--16",
isbn="978-3-540-73424-6",
doi="10.1007/978-3-540-73424-6_2",
url="https://doi.org/10.1007/978-3-540-73424-6_2"
}

@article{TANG2018132,
title = {Using auction-based task allocation scheme for simulation optimization of search and rescue in disaster relief},
journal = {Simulation Modelling Practice and Theory},
volume = {82},
pages = {132-146},
year = {2018},
issn = {1569-190X},
doi = {https://doi.org/10.1016/j.simpat.2017.12.014},
url = {https://www.sciencedirect.com/science/article/pii/S1569190X17301934},
author = {Jian Tang and Kejun Zhu and Haixiang Guo and Chengzhu Gong and Can Liao and Shuwen Zhang}
}

@article{SARSurvey,
author = {Peña Queralta, Jorge and Taipalmaa, Jussi and Pullinen, Bilge and Sarker, Victor and Nguyen gia, Tuan and Tenhunen, Hannu and Gabbouj, Moncef and Raitoharju, Jenni and Westerlund, Tomi},
year = {2020},
month = {01},
pages = {191617-191643},
title = {Collaborative Multi-Robot Search and Rescue: Planning, Coordination, Perception, and Active Vision},
volume = {8},
journal = {IEEE Access},
doi = {10.1109/ACCESS.2020.3030190}
}

@ARTICLE{disttask,
  author={Zhao, Wanqing and Meng, Qinggang and Chung, Paul W. H.},
  journal={IEEE Transactions on Cybernetics}, 
  title={A Heuristic Distributed Task Allocation Method for Multivehicle Multitask Problems and Its Application to Search and Rescue Scenario}, 
  year={2016},
  volume={46},
  number={4},
  pages={902-915},
  doi={10.1109/TCYB.2015.2418052}}

@INPROCEEDINGS{thompsonsearch,
  author={Chen, Yidan and Gao, Melvin Ticiano},
  booktitle={2022 IEEE 2nd International Conference on Data Science and Computer Application (ICDSCA)}, 
  title={Multi-target Search Using Teams of Robots Based on Thompson Sampling}, 
  year={2022},
  volume={},
  number={},
  pages={720-723},
  doi={10.1109/ICDSCA56264.2022.9988679}}

@article{activethompson,
  author  = {Chen, Jun and Dames, Philip},
  title   = {Active Multi-Target Search Using Distributed Thompson Sampling},
  journal = {Research Square},
  year    = {2022},
  month   = jul,
  day     = {15},
  note    = {Preprint, Version 1},
  doi     = {10.21203/rs.3.rs-1849567/v1},
  url     = {https://doi.org/10.21203/rs.3.rs-1849567/v1}
}

@INPROCEEDINGS{distthompson,
  author={Liang, Hongrui and Yan, Bowen and Yang, Runyu and Zhao, Junxiang},
  booktitle={2022 International Symposium on Control Engineering and Robotics (ISCER)}, 
  title={Sampling-based Distributed Multi-target Search Using A Team of Mobile Robots}, 
  year={2022},
  volume={},
  number={},
  pages={76-82},
  doi={10.1109/ISCER55570.2022.00019}}

@mastersthesis{Doke2012AnalysisOS,
  title={Analysis of Search Incidents and Lost Person Behavior in Yosemite National Park},
  author={Jared Doke},
  year={2012},
  school={University of Kansas },
  url={https://api.semanticscholar.org/CorpusID:128032156}
}

@book{Koester2008,
  author    = {Koester, Robert J.},
  title     = {Lost Person Behavior: A Search and Rescue Guide on Where to Look---for Land, Air and Water},
  publisher = {dbS Productions LLC},
  year      = {2008}
}

@article{mapscore,
author = {Sava, Elena and Twardy, Charles and Koester, Robert and Sonwalkar, Mukul},
title = {Evaluating Lost Person Behavior Models},
journal = {Transactions in GIS},
volume = {20},
number = {1},
pages = {38-53},
doi = {https://doi.org/10.1111/tgis.12143},
url = {https://onlinelibrary.wiley.com/doi/abs/10.1111/tgis.12143},
eprint = {https://onlinelibrary.wiley.com/doi/pdf/10.1111/tgis.12143},
year = {2016}
}

@book{Koopman1980,
  author    = {Koopman, Bernard Osgood},
  title     = {Search and Screening: General Principles with Historical Applications},
  year      = {1980},
  note      = {Originally published by Pergamon Press (1946); reprinted edition},
  publisher = {Society for Industrial Applied Mathematics}
}

@INPROCEEDINGS{nats,
  author={Ghods, Ramina and Durkin, William J. and Schneider, Jeff},
  booktitle={2021 IEEE International Conference on Robotics and Automation (ICRA)}, 
  title={Multi-Agent Active Search using Realistic Depth-Aware Noise Model}, 
  year={2021},
  volume={},
  number={},
  pages={9101-9108},
  doi={10.1109/ICRA48506.2021.9561598}}

@InProceedings{ghods,
  title = 	 {Decentralized multi-agent active search for sparse signals},
  author =       {Ghods, Ramina and Banerjee, Arundhati and Schneider, Jeff},
  booktitle = 	 {Proceedings of the Thirty-Seventh Conference on Uncertainty in Artificial Intelligence},
  pages = 	 {696--706},
  year = 	 {2021},
  editor = 	 {de Campos, Cassio and Maathuis, Marloes H.},
  volume = 	 {161},
  series = 	 {Proceedings of Machine Learning Research},
  month = 	 {27--30 Jul},
  publisher =    {PMLR},
  url = 	 {https://proceedings.mlr.press/v161/ghods21a.html}
}

@inproceedings{grontved2022icar,
  title={Decentralized Multi-UAV Trajectory Task Allocation in Search and Rescue Applications},
  author={Gr{\o}ntved, Kasper Andreas R{\o}mer and Schultz, Ulrik Pagh and Christensen, Anders Lyhne},
  booktitle={21st International Conference on Advanced Robotics},
  year={2023},
  organization={IEEE}
}

@article{perceptionuncert,
author = {Yang, Mingliang and Jiang, Kun and Wen, Junze and Peng, Liang and Yang, Yanding and Wang, Hong and Yang, Mengmeng and Jiao, Xinyu and Yang, Diange},
year = {2023},
month = {03},
pages = {2867},
title = {Real-Time Evaluation of Perception Uncertainty and Validity Verification of Autonomous Driving},
volume = {23},
journal = {Sensors},
doi = {10.3390/s23052867}
}

@article{Ewers_2025,
   title={Predictive Probability Density Mapping for Search and Rescue Using an Agent-Based Approach With Sparse Data},
   volume={13},
   ISSN={2169-3536},
   url={http://dx.doi.org/10.1109/ACCESS.2025.3557693},
   DOI={10.1109/access.2025.3557693},
   journal={IEEE Access},
   publisher={Institute of Electrical and Electronics Engineers (IEEE)},
   author={Ewers, Jan-Hendrik and Anderson, David and Thomson, Douglas},
   year={2025},
   pages={60802–60813} }

@article{Ewers_2023,
author = {Ewers, Jan-Hendrik and Anderson, David and Thomson, Douglas},
title = {Optimal path planning using psychological profiling in drone-assisted missing person search},
journal = {Advanced Control for Applications},
volume = {5},
number = {4},
pages = {e167},
doi = {https://doi.org/10.1002/adc2.167},
url = {https://onlinelibrary.wiley.com/doi/abs/10.1002/adc2.167},
eprint = {https://onlinelibrary.wiley.com/doi/pdf/10.1002/adc2.167},
year = {2023}
}

@InProceedings{ghods2021,
  title = 	 {Decentralized multi-agent active search for sparse signals},
  author =       {Ghods, Ramina and Banerjee, Arundhati and Schneider, Jeff},
  booktitle = 	 {Proceedings of the Thirty-Seventh Conference on Uncertainty in Artificial Intelligence},
  pages = 	 {696--706},
  year = 	 {2021},
  editor = 	 {de Campos, Cassio and Maathuis, Marloes H.},
  volume = 	 {161},
  series = 	 {Proceedings of Machine Learning Research},
  month = 	 {27--30 Jul},
  publisher =    {PMLR},
  url = 	 {https://proceedings.mlr.press/v161/ghods21a.html}
}

@ARTICLE{spantree,
  author={Dong, Wei and Liu, Sensen and Ding, Ye and Sheng, Xinjun and Zhu, Xiangyang},
  journal={IEEE Transactions on Automation Science and Engineering}, 
  title={An Artificially Weighted Spanning Tree Coverage Algorithm for Decentralized Flying Robots}, 
  year={2020},
  volume={17},
  number={4},
  pages={1689-1698},
  doi={10.1109/TASE.2020.2971324}}

@ARTICLE{rlcover,
  author={Xiao, Jian and Wang, Gang and Zhang, Ying and Cheng, Lei},
  journal={IEEE Access}, 
  title={A Distributed Multi-Agent Dynamic Area Coverage Algorithm Based on Reinforcement Learning}, 
  year={2020},
  volume={8},
  number={},
  pages={33511-33521},
  doi={10.1109/ACCESS.2020.2967225}}

@ARTICLE{coopplan,
  author={Tang, Yuan and Zhou, Rui and Sun, Guibin and Di, Bin and Xiong, Rongling},
  journal={IEEE Sensors Journal}, 
  title={A Novel Cooperative Path Planning for Multirobot Persistent Coverage in Complex Environments}, 
  year={2020},
  volume={20},
  number={8},
  pages={4485-4495},
  doi={10.1109/JSEN.2019.2963697}}

@INPROCEEDINGS{linuav,
  author={Lin, Lanny and Goodrich, Michael A.},
  booktitle={2009 IEEE/RSJ International Conference on Intelligent Robots and Systems}, 
  title={UAV intelligent path planning for Wilderness Search and Rescue}, 
  year={2009},
  volume={},
  number={},
  pages={709-714},
  doi={10.1109/IROS.2009.5354455}}

@Inbook{MRTA,
author="Khamis, Alaa
and Hussein, Ahmed
and Elmogy, Ahmed",
editor="Koub{\^a}a, Anis
and Mart{\'i}nez-de Dios, J.Ramiro",
title="Multi-robot Task Allocation: A Review of the State-of-the-Art",
bookTitle="Cooperative Robots and Sensor Networks 2015",
year="2015",
publisher="Springer International Publishing",
address="Cham",
pages="31--51",
isbn="978-3-319-18299-5",
doi="10.1007/978-3-319-18299-5_2",
url="https://doi.org/10.1007/978-3-319-18299-5_2"
}

@article{virtualstyg,
title = {Multi-agent systems with virtual stigmergy},
journal = {Science of Computer Programming},
volume = {187},
pages = {102345},
year = {2020},
issn = {0167-6423},
doi = {https://doi.org/10.1016/j.scico.2019.102345},
url = {https://www.sciencedirect.com/science/article/pii/S016764231930139X},
author = {Rocco {De Nicola} and Luca {Di Stefano} and Omar Inverso}
}

\end{document}